\documentclass[11pt,a4paper]{article}
\usepackage[hyperref]{acl2021}
\usepackage{times}
\usepackage{latexsym}

\usepackage{microtype}

\aclfinalcopy 

\usepackage{xcolor}
\definecolor{blue}{HTML}{1F78B4}
\definecolor{green}{HTML}{33A02C}
\definecolor{red}{HTML}{E31A1C}

\usepackage{listings}
\usepackage{array}
\usepackage{graphicx}
\usepackage{booktabs}
\usepackage{multirow}
\usepackage{soul}
\usepackage{makecell}
\usepackage{enumitem}

\title{\texttt{OpenAttack}: An Open-source Textual Adversarial Attack Toolkit}

\author{
Guoyang Zeng$^{1,2}$\thanks{\ \ Indicates equal contribution}\hspace{0.3em},
Fanchao Qi$^{1,2*}$,
Qianrui Zhou$^{1,2}$,
Tingji Zhang$^{1,2}$,
Zixian Ma$^{4}$\thanks{\ \ Work done during internship at Tsinghua University}\hspace{0.3em},
\\
{\bf Bairu Hou$^{2,5}$, Yuan Zang$^{1,2}$, Zhiyuan Liu$^{1,2,3}$\thanks{\ \  Corresponding author. Email: liuzy@tsinghua.edu.cn}\hspace{0.3em},
Maosong Sun$^{1,2,3}$ 
}
\\
$^{1}$Department of Computer Science and Technology, Tsinghua University, Beijing, China \\
$^{2}$Beijing National Research Center for Information Science and Technology\\
$^{3}$Institute for Artificial Intelligence, Tsinghua University, Beijing, China \\
$^{4}$Stanford University\quad 
$^{5}$School of Economics and Management, Tsinghua University \\
{\tt zenggy@mail.tsinghua.edu.cn, qfc17@mails.tsinghua.edu.cn} 
}

\date{}

\begin{document}
\maketitle
\begin{abstract}
Textual adversarial attacking has received wide and increasing attention in recent years.
Various attack models have been proposed, which are enormously distinct and implemented with different programming frameworks and settings.
These facts hinder quick utilization and fair comparison of attack models.
In this paper, we present an open-source textual adversarial attack toolkit named \texttt{OpenAttack} to solve these issues.
Compared with existing other textual adversarial attack toolkits, \texttt{OpenAttack} has its unique strengths in support for all attack types, multilinguality, and parallel processing.
Currently, \texttt{OpenAttack} includes 15 typical attack models that cover all attack types. 
Its highly inclusive modular design not only supports quick utilization of existing attack models, but also enables great flexibility and extensibility.
\texttt{OpenAttack} has broad uses including comparing and evaluating attack models, measuring robustness of a model, assisting in developing new attack models, and adversarial training.
Source code and documentation can be obtained at \url{https://github.com/thunlp/OpenAttack}.

\end{abstract}

\section{Introduction}
Deep neural networks (DNNs) have been found to be susceptible to adversarial attacks \citep{szegedy2014intriguing,goodfellow2015explaining}.
The attacker uses adversarial examples, which are maliciously crafted by imposing small perturbations on original input, to fool the victim model.
With the wide application of DNNs to practical systems accompanied by growing concern about their security, research on adversarial attacking has become increasingly important.
Moreover, adversarial attacks are also helpful to improve robustness and interpretability of DNNs \citep{wallace2019universal}.

In the field of natural language processing (NLP), diverse adversarial attack models have been proposed \citep{zhang2020adversarial}.
These models vary in \textit{accessibility} to the victim model (ranging from having full knowledge to total ignorance) and \textit{perturbation level} (character-, word- or sentence-level).
In addition, they are originally proposed to attack different victim models on different NLP tasks under different evaluation protocols.

This immense diversity causes serious difficulty for fair and apt comparison between different attack models, which is unfavourable to the development of textual adversarial attacking.
Further, although most attack models are open-source, they use different programming frameworks and settings, which lead to unnecessary time and effort when implementing them.

\begin{table*}[t]
\centering
\resizebox{\linewidth}{!}{
\begin{tabular}{l|cc|l}
    \toprule
    \multicolumn{1}{c|}{Model} & Accessibility & Perturbation & \multicolumn{1}{c}{Main Idea} \\
    \midrule
    SEA \citep{ribeiro2018semantically} & Decision& Sentence & Rule-based paraphrasing\\ 
    SCPN \citep{iyyer2018adversarial} & Blind & Sentence & Paraphrasing \\ 
    GAN \citep{zhao2018generating} & Decision & Sentence & Text generation by encoder-decoder\\ 
    TextFooler \citep{jin2020bert} & Score & Word & Greedy word substitution \\ 
    PWWS \citep{ren2019generating} & Score & Word & Greedy word substitution  \\
    Genetic \citep{alzantot2018generating} & Score & Word & Genetic algorithm-based word substitution  \\ 
    SememePSO \citep{zang2020word} & Score & Word & Particle swarm optimization-based word substitution  \\ 
    BERT-ATTACK \citep{li2020bert} & Score & Word & Greedy contextualized word substitution  \\ 
    BAE \citep{garg2020bae} & Score & Word & Greedy contextualized word substitution and insertion  \\ 
    FD \citep{papernot2016crafting} & Gradient & Word & Gradient-based word substitution  \\ 
    TextBugger \citep{li2019textbugger} & Gradient, Score & Word+Char & Greedy word substitution and character manipulation \\ 
     UAT \citep{wallace2019universal} & Gradient & Word, Char & Gradient-based word or character manipulation\\ 
     HotFlip \citep{ebrahimi2018hotflip} & Gradient & Word, Char & Gradient-based word or character substitution \\
     VIPER \citep{eger2019text} & Blind & Char & Visually similar character substitution \\ 
     DeepWordBug \citep{gao2018black} & Score & Char & Greedy character manipulation\\ 
    \bottomrule
\end{tabular}
}
\caption{Textual adversarial attack models involved in \texttt{OpenAttack}, among which the three sentence-level models SEA, SCPN and GAN together with FD, UAT and VIPER are not included in TextAttack for now. ``Accessibility'' is the accessibility to the victim model, and ``Perturbation'' refers to perturbation level. 
``Sentence'', ``Word'' and ``Char'' denote sentence-, word- and character-level perturbations.
In the columns of Accessibility and Perturbation, ``A, B'' means that the attack model supports both A and B , while ``A+B'' means that the attack model conducts A and B simultaneously.}
\label{tab:attack_models}
\end{table*}

To tackle these challenges, a textual adversarial attacking toolkit named TextAttack \citep{morris2020textattack} has been developed. 
It implements several textual adversarial attack models under a unified framework and provides interfaces for utilizing existing attack models or designing new attack models. 
So far, TextAttack has attracted considerable attention and facilitated the birth of new attack models such as BAE \citep{garg2020bae}.

In this paper, we present \texttt{OpenAttack}, which is also an open-source toolkit for textual adversarial attacking.
Similar to TextAttack, \texttt{OpenAttack} adopts modular design to assemble various attack models, in order to enable quick implementation of existing or new attack models.
But \texttt{OpenAttack} is different from and complementary to TextAttack mainly in the following three aspects:

(1) \textbf{Support for all attacks}. TextAttack utilizes a relatively rigorous framework to unify different attack models. However, this framework is naturally not suitable for sentence-level adversarial attacks, an important and typical kind of textual adversarial attacks. 
Thus, no sentence-level attack models are included in TextAttack. 
In contrast, \texttt{OpenAttack} adopts a more flexible framework that supports all types of attacks including sentence-level attacks.

(2) \textbf{Multilinguality}. TextAttack only covers English textual attacks while \texttt{OpenAttack} supports English and Chinese now. And its extensible design enables quick support for more languages.

(3) \textbf{Parallel processing}. Running some attack models maybe very time-consuming, e.g., it takes over 100 seconds to attack an instance with the SememePSO attack model \citep{zang2020word}.
To address this issue, \texttt{OpenAttack} additionally provides support for multi-process running of attack models to improve attack efficiency.

Moreover, \texttt{OpenAttack} is fully integrated with HuggingFace's \textsl{transformers}\footnote{\url{https://github.com/huggingface/transformers}} and \textsl{datasets}\footnote{\url{https://github.com/huggingface/datasets}} libraries, which allows convenient adversarial attacks against thousands of NLP models (especially pre-trained models) on diverse datasets.
\texttt{OpenAttack} also has great extensibility. It can be easily used to attack any customized victim model, regardless of the used programming framework (PyTorch, TensorFlow, Keras, etc.), on any customized dataset.


\texttt{OpenAttack} can be used to (1) provide various handy baselines for attack models; (2) comprehensively evaluate attack models using its thorough evaluation metrics; (3) assist in quick development of new attack models; (4) evaluate the robustness of an NLP model against various adversarial attacks; and (5) conduct \textit{adversarial training} \citep{goodfellow2015explaining} to improve model robustness by enriching the training data with generated adversarial examples.

Recent years have witnesses the rapid development of adversarial attacks in computer vision \citep{akhtar2018threat}, which is promoted by many visual attack toolkits such as CleverHans \citep{papernot2018cleverhans}, Foolbox \citep{rauber2017foolbox}, AdvBox \citep{goodman2020advbox}, etc.
We hope \texttt{OpenAttack}, together with TextAttack and other similar toolkits, can play a constructive role in the development of textual adversarial attacks.

\section{Formalization and Categorization of Textual Adversarial Attacking}
We first formalize the task of textual adversarial attacking for text classification, and the following formalization can be trivially adapted to other NLP tasks.
For a given text sequence $x$ that is correctly classified as its ground-truth label $y$ by the victim model $F$, the attack model $A$ is supposed to transform $x$ into $\hat{x}$ by small perturbations, whose ground-truth label is still $y$ but classification result given by $F$ is $\hat{y}\neq y$.
Next, we introduce the categorization of textual adversarial attack models from three perspectives.

According to the attack model's accessibility to the victim model, existing attack models can be categorized into four classes, namely gradient-based, score-based, decision-based and blind models.
First, gradient-based attack models are also called white-box attack models, which require full knowledge of the victim model to conduct gradient update. Most of them are inspired by the fast gradient sign method \citep{goodfellow2015explaining} and forward derivative method \citep{papernot2016limitations} in visual adversarial attacking.

In contrast to white-box attack models, black-box models do not need to have complete information on the victim model, and can be subcategorized into score-based, decision-based and blind models.
Blind models are ignorant of the victim model at all. 
Score-based models require the prediction scores (e.g., classification probabilities) of the victim model, while decision-based models only need the final decision (e.g., predicted class).

According to the level of perturbations imposed on original input, textual adversarial attack models can be classified into sentence-level, word-level and character-level models.
Sentence-level attack models craft adversarial examples mainly by adding distracting sentences \citep{jia2017adversarial}, paraphrasing \citep{iyyer2018adversarial,ribeiro2018semantically} or text generation by encoder-decoder \citep{zhao2018generating}.
Word-level attack models mostly conduct word substitution, namely substituting some words in the original input with semantically identical or similar words such as synonyms \citep{jin2020bert,ren2019generating,alzantot2018generating}.
Some word-level attack models also use operations including deleting and adding words \citep{zhang2019generating,garg2020bae}.
Character-level attack models usually carry out character manipulations including swap, substitution, deletion, insertion and repeating \citep{eger2019text,ebrahimi2018hotflip,belinkov2018synthetic}.

Finally, adversarial attack models can also be categorized into targeted and untargeted models based on whether the wrong classification result given by the victim model ($\hat{y}$) is pre-specified (mainly for the multi-class classification models).
Most existing attack models support (by minor adjustment) both targeted and untargeted attacks, and we give no particular attention to this attribute of attack models in this paper.

Currently \texttt{OpenAttack} includes $15$ different attack models, which cover all the victim model accessibility and perturbation level types.
Table \ref{tab:attack_models} lists the attack models involved in \texttt{OpenAttack}.

\begin{figure}[!t]
\setlength{\belowcaptionskip}{-5pt}   
    \centering
    \includegraphics[width=\linewidth]{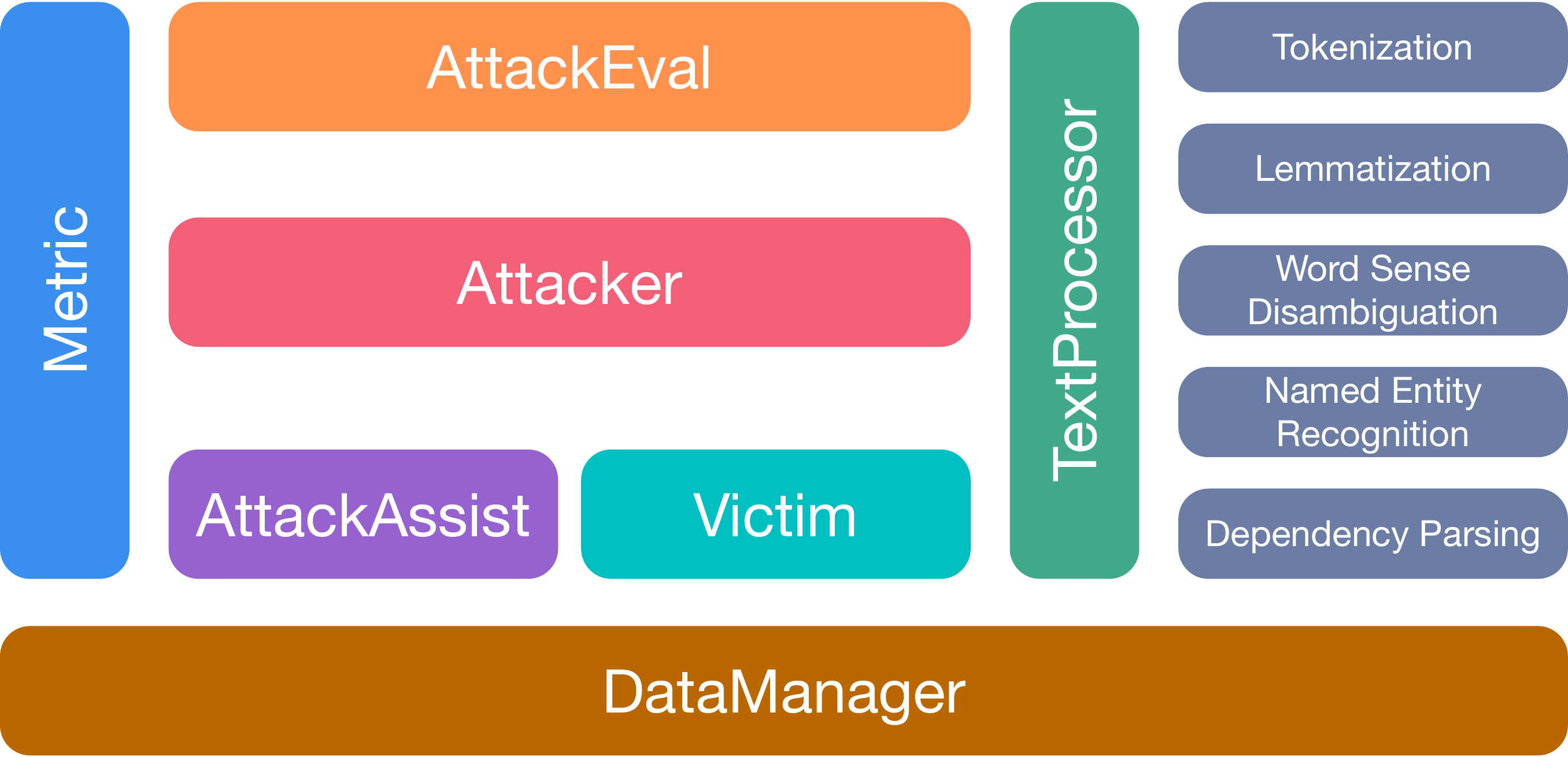}
    \caption{Overall architecture of \texttt{OpenAttack}.}
    \label{fig:frame}
\end{figure}

\section{Toolkit Design and Architecture}
In this section, we describe the design philosophy and modular architecture of \texttt{OpenAttack}.

We extract and properly reorganize the commonly used components from different attack models, so that any attack model can be assembled by them.
Considering the significant distinctions among different attack models, especially those between the sentence-level and word/char-level attack models, it is hard to embrace all attack models within a unified framework like TextAttack.
Therefore, we leave considerable freedom for the skeleton design of attack models, and focus more on streamlining the general processing of adversarial attacking and providing common components used in attack models.
Next we introduce the modules of \texttt{OpenAttack} one by one, and Figure \ref{fig:frame} illustrates an overview of all the modules.

\begin{itemize}
 [leftmargin=12pt]
    \item \textbf{TextProcessor}. This module is aimed at processing the original input so as to assist attack models in generating adversarial examples. It consists of several functions used for tokenization, lemmatization, delemmatization, word sense disambiguation (WSD), named entity recognition (NER) and dependency parsing. Currently it supports English and Chinese, and support for other languages can be realized simply by rewriting the TextProcessor base class.
    \item \textbf{Victim}. This module wraps the victim model. It supports both neural network-based model implemented by any programming framework (especially the HuggingFace's \textsl{transformers}) and traditional machine learning model. It is mainly composed of three functions that are used to obtain the gradient with respect to the input, prediction scores and predicted class of a victim model.
    
    \item \textbf{Attacker}. This is the core module of \texttt{OpenAttack}. It comprises various attack models and can generate adversarial examples for given input against a victim model.
    
    \item \textbf{AttackAssist}. This is an assistant module of Attacker. It mainly packs different word and character substitution methods that are widely used in word- and character-level attack models. Attacker queries this module to get substitutions for a word or character. 
    Now it includes word embedding-based \citep{alzantot2018generating,jin2020bert}, synonym-based \citep{ren2019generating} and sememe-based \citep{zang2020word} word substitution methods, and visual character substitution method \citep{eger2019text}.
    In addition, some useful components used in sentence-level attack models are also included, such as paraphrasing based on back-translation.

    \item \textbf{Metric}. This module provides several adversarial example quality metrics which can serve as either the constraints on the adversarial examples during attacking or evaluation metrics for evaluating adversarial attacks. It currently includes following metrics: (1) language model prediction score for a given word in a context given by Google one-billion words language model \citep{jozefowicz2016exploring} (this metric can be used as the constraint on adversarial examples only); (2) \textit{word modification rate}, the percentage of modified words of an adversarial example compared with the original example; (3) formal similarity between the adversarial example and original example, which is measured by Levenshtein edit distance \citep{levenshtein1966binary}, character- and word-level Jaccard similarity \citep{jaccard1912} and BLEU score \citep{papineni2002bleu}; (4) semantic similarity between the adversarial example and original example measured by Universal Sentence Encoder \citep{cer2018universal} and Sentence Transformers \citep{reimers2019sentence}; (5) adversarial example fluency measured with perplexity computed by GPT-2 \citep{radford2019language}; and (6) grammaticality measure by the grammatical errors given by LanguageTool.\footnote{\url{https://www.languagetool.org}} 
    
\begin{table}[t]
    \centering
    \resizebox{\linewidth}{!}{
    \begin{tabular}{c|c|c}
    \toprule
        Perspective & Metric & Better? \\
        \midrule
       \makecell[c]{Attack \\Effectiveness}  & Attack Success Rate & Higher \\
       \hline
       \multirow{5}{*}{\makecell[c]{Adversarial \\Example \\Quality}}  & Word Modification Rate & Lower \\
         & Formal Similarity & Higher  \\
         & Semantic Similarity & Higher \\
         & Fluency (GPT-2 perplexity) & Lower \\
         & Grammaticality (Grammatical Errors)& Lower  \\
       \hline
       \multirow{2}{*}{\makecell[c]{Attack \\Efficiency}}  & Average Victim Model Query Times & Lower \\
         & Average Running Time & Lower \\
         \bottomrule
    \end{tabular}
    }
    \caption{Evaluation metrics in \texttt{OpenAttack}. ``Higher'' and ``Lower'' mean the higher/lower the metric is, the better an attack model performs.}
    \label{tab:metrics}
\end{table}
    \item \textbf{AttackEval}. This module is used to evaluate textual adversarial attacks from different perspectives including attack effectiveness, adversarial example quality and attack efficiency: 
    (1) the attack effectiveness metric is \textit{attack success rate}, the percentage of the attacks that successfully fool the victim model;
    (2) adversarial example quality is measured by the last five metrics in the Metric module; and
    (3) attack efficiency has two metrics including average victim model query times and average running time of attacking one instance.
    Table \ref{tab:metrics} lists all the evaluation metrics in \texttt{OpenAttack}.
    
    The realization of multi-processing is incorporated in this module, with the help of Python \texttt{multiprocessing} library.
    In addition, this module can also visualize and save attack results, e.g., display original input and  adversarial examples and emphasize their differences.
    
    \item \textbf{DataManager}. This module manages all the data as well as saved models that are used in other modules. It supports accessing and downloading data/models.
    Specifically, it deals with the data used in the AttackAssist module such as character embeddings, word embeddings and WordNet synonyms, the models used in the TextProcessor module such as NER model and dependency parser, the built-in trained victim models, and auxiliary models used in Attacker such as the paraphrasing model for the paraphrasing-based attack models. This module helps efficiently and handily utilize data.
\end{itemize}

\section{Toolkit Usage}
\texttt{OpenAttack} provides a set of easy-to-use interfaces that can meet almost all the needs in textual adversarial attacking, such as preprocessing text, generating adversarial examples to attack a victim model and evaluating attack models.
Moreover, \texttt{OpenAttack} has great flexibility and extensibility and supports easy customization of victim models and attack models.
Next, we showcase some basic usages of \texttt{OpenAttack}.

\subsection{Built-in Attack and Evaluation}
\texttt{OpenAttack} builds in some commonly used NLP models such as LSTM \citep{hochreiter1997long} and BERT \citep{devlin2019bert} that have been trained on commonly used NLP datasets.
Users can use the built-in victim models to quickly conduct adversarial attacks.
The following code snippet shows how to use Genetic \citep{alzantot2018generating} to attack BERT on the test set of SST-2 \citep{socher2013recursive} with 4-process parallel running:

\begin{lstlisting}
import OpenAttack as oa
import datasets # HuggingFace's datasets library
import multiprocessing 
# choose a trained victim model
victim = oa.loadVictim('BERT.SST')
# choose a evaluation dataset from datasets
dataset = datasets.load_dataset('sst', 'test')
# choose Genetic as the attacker
attacker = oa.attackers.GeneticAttacker() 
# prepare for multi-process attacking
attack_eval = oa.attack_evals.MultiProcessAttackEval(attacker, victim, num_process=4)
# launch attacks and print attack results 
attack_eval.eval(dataset, visualize=True)
\end{lstlisting}

\begin{figure}[!t]
    \centering
    \includegraphics[width=\linewidth]{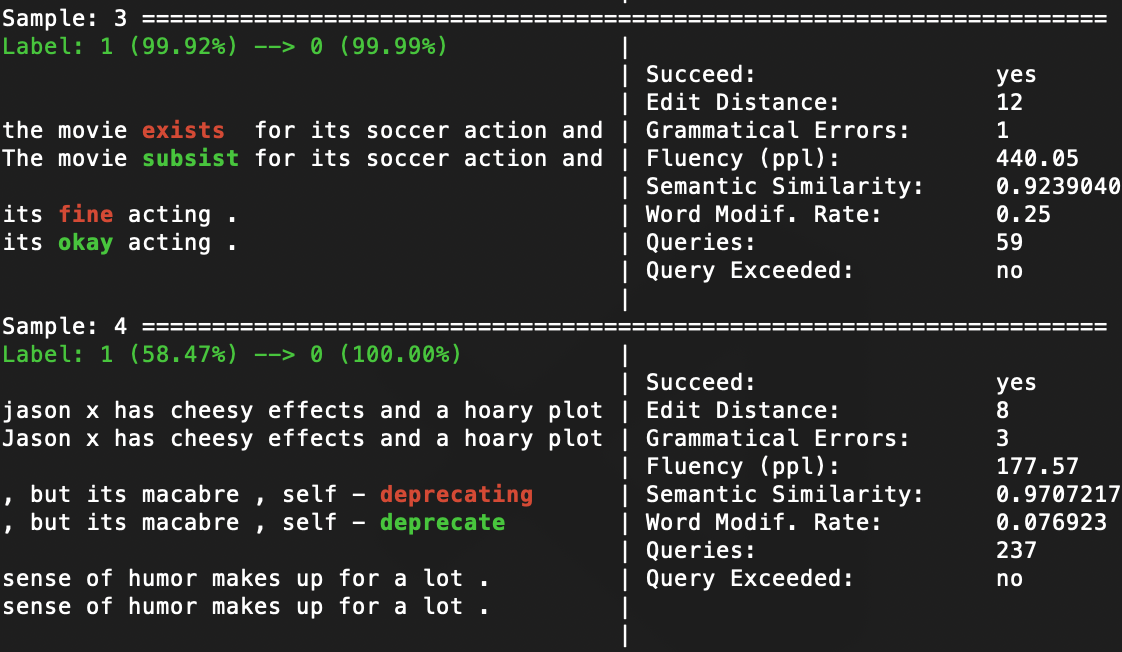}
    \caption{Part of attack results for individual instances.}
    \label{fig:result_each}
\end{figure}

\begin{figure}[!t]
    \centering
    \includegraphics[width=\linewidth]{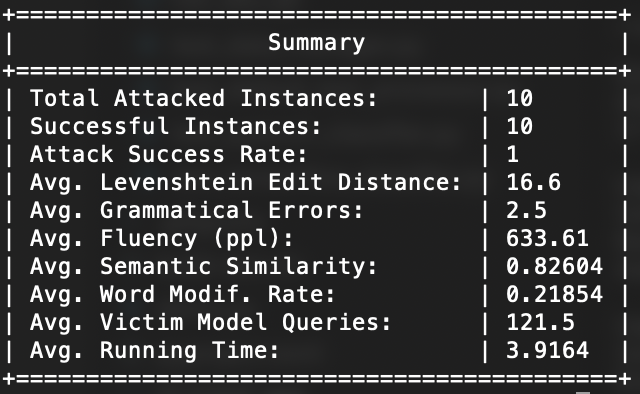}
    \caption{Summary of attack results.}
    \label{fig:result_sum}
\end{figure}

Figure \ref{fig:result_each} displays the printed attack results for individual instances from above code.
We can see \texttt{OpenAttack} prints the original input as well as the word-aligned adversarial example, where the perturbed words are colored.
In addition, a series of attack evaluation results are listed.
At the end of individual attack results, \texttt{OpenAttack} provides an attack summary composed of average evaluation results of specified metrics among all instances, as shown in Figure \ref{fig:result_sum}.

\begin{table*}[t]
\centering
\resizebox{\linewidth}{!}{
\begin{tabular}{c||cc||c|crccr|rrrr}
    \toprule
    \multirow{2}{*}{Model} & \multicolumn{2}{c||}{Type} & Effectiveness & \multicolumn{5}{c|}{Adversarial Example Quality}  & \multicolumn{4}{c}{Attack Efficiency} \\
    \cmidrule{2-13}
    &Accessibility & Perturbation & ASR & WMR & LES & SemSim & Fluency & Grm & \#Query & T$_1$ & T$_2$ & S \\
    \midrule
    SEA  &  Decision & Sentence &  0.12 & -- & 14.7 & 0.90 & 398 & 2.2 & 2.0 & 37.1 & -- & --  \\
    SCPN & Blind & Sentence & 0.68 & -- & 55.6 & 0.56 & 432 & 2.7 & 11.0 & 3.58 & 2.30 & 1.56 \\
    GAN & Decision & Sentence &  0.41 & -- & 68.8 & 0.26 & 512 & 4.2 & 2.0 & 0.60 & 2.00 & 0.30 \\
    TextFooler & Score & Word  & 0.90 & 0.11 & 14.1 & 0.87 & 621 & 4.6 & 130.5 & 5.75 & 3.25 & 1.77 \\
    PWWS  & Score & Word & 0.78 & 0.20 & 17.9 & 0.84 & 613 & 2.9 & 124.8 & 5.26 & 2.88 & 1.83  \\
    Genetic & Score & Word &  0.36 & 0.11 & 13.4 & 0.88 & 689 & 4.7 & 242.1 & 54.11 & 27.56 & 1.96 \\
    SememePSO & Score & Word &   0.82 &  0.14 & 2.9 & 0.89  & 711  & 2.9  &  177.9 &   102.44 & 52.41 & 1.95 \\
    BERT-ATTACK & Score & Word & 0.87 & 0.31 & 4.2 & 0.86 & 796& 4.4 & 51.9 & 2.38 & 1.57 & 1.51 \\
    BAE & Score & Word & 0.77 &  0.68 & 5.4 & 0.82  & 1147  & 4.3  & 103.0 &    2.97 & 1.79 & 1.66 \\
    FD  & Gradient & Word &  0.16 & 0.24 & 17.9 & 0.85 & 908 & 3.1 & 10.9 & 34.57 & 28.36 & 1.22 \\ 
    TextBugger & Gradient & Word+Char &  0.25  & 0.15 & 10.6 & 0.61 & 512 & 7.1 & 150.0 & 8.49 & 4.37 & 1.94 \\
     UAT  & Gradient & Word & 0.43 & 0.15 & 24.0 & 0.85 & 620 & 2.8 & 2.0 & 0.08 & -- & --  \\
     HotFlip & Gradient & Word  &  0.47 & 0.08 & 8.9 & 0.93 & 333 & 2.7 & 105.4 & 2.77 & 1.82 & 1.52 \\
     VIPER  & Blind & Char &  0.27 & -- & 24.2 & 0.22 & 347 & 15.8 & 3.0 & 4.01 & 2.04 & 1.97\\
     DeepWordBug  & Score & Char & 0.46 & -- & 7.9 & 0.73 & 731 & 6.1 & 22.0 & 0.97 & 0.62 & 1.56 \\
    \bottomrule
\end{tabular}
}
\caption{Evaluation results of different attack models when attacking BERT on SST-2. ASR = Attack Success Rate, WMR = Word Modification Rate that is only applicable to word-level attacks, LES = Levenshterin Edit Distance, SemSim = Semantic Similarity measured by Universal Sentence Encoder, Fluency = GPT-2 perplexity, Grm = number of grammatical errors, \#Query = average victim model query times, T$_1$ and T$_2$ represent average running time of attacking one instance (seconds) with single and dual process, S = T$_1$/T$_2$ is speedup. Notice that it is meaningless to run SEA and UAT with multi-process because they learn and conduct global perturbations. 
}
\label{tab:eval}
\end{table*}

\subsection{Customized Victim Models}
It is very common for users to launch attacks against their own models that have been trained on specific datasets, particularly when evaluating the robustness of a victim model.
It is impossible to exhaustively build in all victim models.
Thus, easy customization for victim models is very important.

\texttt{OpenAttack} provides simple and convenient interfaces for victim model customization.
For a trained model implemented with whichever programming framework, users just need to configure some model access interfaces that provide accessibility required for the attack model under the Victim class.
The following code snippet shows how to use Genetic to attack a customized sentiment analysis model, a statistical model in NLTK \citep{bird2009natural}, on the test set of SST.

\begin{lstlisting}
import OpenAttack as oa
import numpy as np
import datasets
from nltk.sentiment.vader import SentimentIntensityAnalyzer

# configure access interface of customized model
class MyModel(oa.Victim):
    def __init__(self):
        self.model = SentimentIntensityAnalyzer()
    def get_prob(self, input_):
        rt = []
        for sent in input_:
            rs = self.model.polarity_scores(sent)
            prob = rs["pos"] / (rs["neg"] + rs["pos"])
            rt.append(np.array([1 - prob, prob]))
        return np.array(rt)
# choose evaluation dataset 
dataset = datasets.load_dataset('sst','test')
# choose the customized victim model
victim = MyModel()
# choose Genetic as the attack model 
attacker = oa.attackers.GeneticAttacker()
# prepare for attacking
attack_eval = oa.attack_evals.DefaultAttackEval(attacker, victim)
# launch attacks and print attack results 
attack_eval.eval(dataset, visualize=True)
\end{lstlisting}

In addition, \texttt{OpenAttack} supports easy customization of attack models thanks to its inclusive modular design.
Due to limited space, please visit the GitHub project page for more examples including attacking HuggingFace's pre-trained models, using customized evaluation metrics and conducting adversarial training.\footnote{\url{https://github.com/thunlp/OpenAttack/tree/master/examples}}

\section{Evaluation}
In this section, we conduct evaluations for all the attack models included in \texttt{OpenAttack}.

We use SST-2 as the evaluation dataset and choose BERT, specifically BERT$_{\rm BASE}$, as the victim model.
After fine-tuning on the training set, BERT achieves 90.31 accuracy on the test set.
Due to great diversity of attack models, it is hard to impose many constraints on attacks like previous work that focuses on a specific kind of attack.
We only restrict the maximum victim model query times to 500.
In addition, to improve evaluation efficiency, we randomly sample $1,000$ correctly classified instances from the test set as the original input to be perturbed.
We use the original default hyper-parameter settings of all attack models.

Table \ref{tab:eval} shows the evaluation results.
By comparison with originally reported results, we confirm the correctness of our implementation. 
We also observe that multi-processing can effectively improve attack efficiency of most attack models (the speedup is greater than 1).
For some very efficient attack models whose average running time is quite short (like GAN), the additional time cost from multi-processing may reduce efficiency instead.


\section{Related Work}
There have been quite a few open-source libraries of generating adversarial examples for continuous data, especially images, such as CleverHans \citep{papernot2018cleverhans}, Foolbox \citep{rauber2017foolbox}, Adversarial Robustness Toolbox (ART) \citep{nicolae2018adversarial} and AdvBox \citep{goodman2020advbox}.
These libraries enable practitioners to easily make adversarial attacks with different methods and have greatly facilitated the development of adversarial attacking for continuous data.

As for discrete data, particularly text, there exist few adversarial attack libraries.
As far as we know, TextAttack \citep{morris2020textattack} is the only such library.
It utilizes a relatively rigorous framework to unify many attack models and provides interfaces for using the existing attack models or designing new attack models. 
As mentioned in Introduction, our \texttt{OpenAttack} is mainly different from and complementary to TextAttack in all-attack-type support, multilinguality and parallel processing.


There are also some other toolkits concerned with textual adversarial attacking. 
TEAPOT \citep{michel2019evaluation} is an open-source toolkit to evaluate the effectiveness of textual adversarial examples from the perspective of preservation of meaning.
It is mainly designed for the attacks against sequence-to-sequence models, but can also be geared towards text classification models.
AllenNLP Interpret \citep{wallace2019allennlp} is a framework for explaining the predictions of NLP models, where adversarial attacking is one of its interpretation methods.
It focuses on interpretability of NLP models and only incorporates two attack models. 


\section{Conclusion and Future Work}
In this paper, we present \texttt{OpenAttack}, an open-source textual adversarial attack toolkit that provides a wide range of functions in textual adversarial attacking.
It is a great complement to existing counterparts because of its unique strengths in all-attack-type support, multilinguality and parallel processing.
Moreover, it has great flexibility and extensibility and provides easy customization of victim models and attack models.
In the future, we will keep \texttt{OpenAttack} updated to incorporate more up-to-date attack models and support more functions to facilitate the research on textual adversarial attacks.

\section*{Acknowledgements}
This work is supported by the National Key Research and Development Program of China (Grant No. 2020AAA0106501 and No. 2020AAA0106502) and Beijing Academy of Artificial Intelligence (BAAI).
We also thank all the anonymous reviewers for their valuable comments and suggestions.

\section*{Broader Impact Statement}
There is indeed a probability that \texttt{OpenAttack} is misused to maliciously attack some NLP systems.
But we believe that we should face up to the potential risks of adversarial attacks rather than pretend not to notice them.
As the development of adversarial learning in computer vision, the studies on adversarial attacks actually promote the studies on adversarial defenses and model robustness.
We hope more people in the NLP community can realize the robustness issue and \texttt{OpenAttack} can play a constructive role.

\bibliographystyle{acl_natbib}
\bibliography{acl2021}

\begin{thebibliography}{41}
\expandafter\ifx\csname natexlab\endcsname\relax\def\natexlab#1{#1}\fi

\bibitem[{Akhtar and Mian(2018)}]{akhtar2018threat}
Naveed Akhtar and Ajmal Mian. 2018.
\newblock Threat of adversarial attacks on deep learning in computer vision: A
  survey.
\newblock \emph{Ieee Access}, 6:14410--14430.

\bibitem[{Alzantot et~al.(2018)Alzantot, Sharma, Elgohary, Ho, Srivastava, and
  Chang}]{alzantot2018generating}
Moustafa Alzantot, Yash Sharma, Ahmed Elgohary, Bo-Jhang Ho, Mani Srivastava,
  and Kai-Wei Chang. 2018.
\newblock Generating natural language adversarial examples.
\newblock In \emph{Proceedings of the EMNLP}.

\bibitem[{Belinkov and Bisk(2018)}]{belinkov2018synthetic}
Yonatan Belinkov and Yonatan Bisk. 2018.
\newblock Synthetic and natural noise both break neural machine translation.
\newblock In \emph{Proceedings of ICLR}.

\bibitem[{Bird et~al.(2009)Bird, Klein, and Loper}]{bird2009natural}
Steven Bird, Ewan Klein, and Edward Loper. 2009.
\newblock \emph{Natural language processing with Python: analyzing text with
  the natural language toolkit}.
\newblock O'Reilly Media, Inc.

\bibitem[{Cer et~al.(2018)Cer, Yang, Kong, Hua, Limtiaco, John, Constant,
  Guajardo-Cespedes, Yuan, Tar et~al.}]{cer2018universal}
Daniel Cer, Yinfei Yang, Sheng-yi Kong, Nan Hua, Nicole Limtiaco, Rhomni~St
  John, Noah Constant, Mario Guajardo-Cespedes, Steve Yuan, Chris Tar, et~al.
  2018.
\newblock Universal sentence encoder for english.
\newblock In \emph{Proceedings of EMNLP}.

\bibitem[{Devlin et~al.(2019)Devlin, Chang, Lee, and
  Toutanova}]{devlin2019bert}
Jacob Devlin, Ming-Wei Chang, Kenton Lee, and Kristina Toutanova. 2019.
\newblock {BERT}: Pre-training of deep bidirectional {Transformers} for
  language understanding.
\newblock In \emph{Proceedings of NAACL-HLT}.

\bibitem[{Ebrahimi et~al.(2018)Ebrahimi, Rao, Lowd, and
  Dou}]{ebrahimi2018hotflip}
Javid Ebrahimi, Anyi Rao, Daniel Lowd, and Dejing Dou. 2018.
\newblock Hotflip: White-box adversarial examples for text classification.
\newblock In \emph{Proceedings of ACL}.

\bibitem[{Eger et~al.(2019)Eger, {\c{S}}ahin, R{\"u}ckl{\'e}, Lee, Schulz,
  Mesgar, Swarnkar, Simpson, and Gurevych}]{eger2019text}
Steffen Eger, G{\"o}zde~G{\"u}l {\c{S}}ahin, Andreas R{\"u}ckl{\'e}, Ji-Ung
  Lee, Claudia Schulz, Mohsen Mesgar, Krishnkant Swarnkar, Edwin Simpson, and
  Iryna Gurevych. 2019.
\newblock Text processing like humans do: Visually attacking and shielding
  {NLP} systems.
\newblock In \emph{Proceedings of NAACL-HLT}.

\bibitem[{Gao et~al.(2018)Gao, Lanchantin, Soffa, and Qi}]{gao2018black}
Ji~Gao, Jack Lanchantin, Mary~Lou Soffa, and Yanjun Qi. 2018.
\newblock Black-box generation of adversarial text sequences to evade deep
  learning classifiers.
\newblock In \emph{Proceedings of IEEE Security and Privacy Workshops}.

\bibitem[{Garg and Ramakrishnan(2020)}]{garg2020bae}
Siddhant Garg and Goutham Ramakrishnan. 2020.
\newblock Bae: Bert-based adversarial examples for text classification.
\newblock In \emph{Proceedings of EMNLP}.

\bibitem[{Goodfellow et~al.(2015)Goodfellow, Shlens, and
  Szegedy}]{goodfellow2015explaining}
Ian~J Goodfellow, Jonathon Shlens, and Christian Szegedy. 2015.
\newblock Explaining and harnessing adversarial examples.
\newblock In \emph{Proceedings of ICLR}.

\bibitem[{Goodman et~al.(2020)Goodman, Xin, Yang, Yuesheng, Junfeng, and
  Huan}]{goodman2020advbox}
Dou Goodman, Hao Xin, Wang Yang, Wu~Yuesheng, Xiong Junfeng, and Zhang Huan.
  2020.
\newblock Advbox: a toolbox to generate adversarial examples that fool neural
  networks.
\newblock \emph{arXiv preprint arXiv:2001.05574}.

\bibitem[{Hochreiter and Schmidhuber(1997)}]{hochreiter1997long}
Sepp Hochreiter and J{\"u}rgen Schmidhuber. 1997.
\newblock Long short-term memory.
\newblock \emph{Neural Computation}, 9(8):1735--1780.

\bibitem[{Iyyer et~al.(2018)Iyyer, Wieting, Gimpel, and
  Zettlemoyer}]{iyyer2018adversarial}
Mohit Iyyer, John Wieting, Kevin Gimpel, and Luke Zettlemoyer. 2018.
\newblock Adversarial example generation with syntactically controlled
  paraphrase networks.
\newblock In \emph{Proceedings of the NAACL-HLT}.

\bibitem[{Jaccard(1912)}]{jaccard1912}
Paul Jaccard. 1912.
\newblock The distribution of the flora in the alpine zone.1.
\newblock \emph{New Phytologist}, 11(2):37--50.

\bibitem[{Jia and Liang(2017)}]{jia2017adversarial}
Robin Jia and Percy Liang. 2017.
\newblock Adversarial examples for evaluating reading comprehension systems.
\newblock In \emph{Proceedings of EMNLP}.

\bibitem[{Jin et~al.(2020)Jin, Jin, Zhou, and Szolovits}]{jin2020bert}
Di~Jin, Zhijing Jin, Joey~Tianyi Zhou, and Peter Szolovits. 2020.
\newblock Is {BERT} really robust? {N}atural language attack on text
  classification and entailment.
\newblock In \emph{Proceedings of AAAI}.

\bibitem[{Jozefowicz et~al.(2016)Jozefowicz, Vinyals, Schuster, Shazeer, and
  Wu}]{jozefowicz2016exploring}
Rafal Jozefowicz, Oriol Vinyals, Mike Schuster, Noam Shazeer, and Yonghui Wu.
  2016.
\newblock Exploring the limits of language modeling.
\newblock \emph{arXiv preprint arXiv:1602.02410}.

\bibitem[{Levenshtein(1966)}]{levenshtein1966binary}
Vladimir~I Levenshtein. 1966.
\newblock Binary codes capable of correcting deletions, insertions, and
  reversals.
\newblock In \emph{Soviet physics doklady}, volume~10, pages 707--710.

\bibitem[{Li et~al.(2019)Li, Ji, Du, Li, and Wang}]{li2019textbugger}
Jinfeng Li, Shouling Ji, Tianyu Du, Bo~Li, and Ting Wang. 2019.
\newblock Textbugger: Generating adversarial text against real-world
  applications.
\newblock In \emph{Proceedings of Network and Distributed Systems Security
  Symposium}.

\bibitem[{Li et~al.(2020)Li, Ma, Guo, Xue, and Qiu}]{li2020bert}
Linyang Li, Ruotian Ma, Qipeng Guo, Xiangyang Xue, and Xipeng Qiu. 2020.
\newblock Bert-attack: Adversarial attack against bert using bert.
\newblock In \emph{Proceedings of EMNLP}.

\bibitem[{Michel et~al.(2019)Michel, Li, Neubig, and
  Pino}]{michel2019evaluation}
Paul Michel, Xian Li, Graham Neubig, and Juan Pino. 2019.
\newblock On evaluation of adversarial perturbations for sequence-to-sequence
  models.
\newblock In \emph{Proceedings NAACL-HLT}.

\bibitem[{Morris et~al.(2020)Morris, Lifland, Yoo, Grigsby, Jin, and
  Qi}]{morris2020textattack}
John Morris, Eli Lifland, Jin~Yong Yoo, Jake Grigsby, Di~Jin, and Yanjun Qi.
  2020.
\newblock Textattack: A framework for adversarial attacks, data augmentation,
  and adversarial training in nlp.
\newblock In \emph{Proceedings of EMNLP: System Demonstrations}.

\bibitem[{Nicolae et~al.(2018)Nicolae, Sinn, Tran, Buesser, Rawat, Wistuba,
  Zantedeschi, Baracaldo, Chen, Ludwig et~al.}]{nicolae2018adversarial}
Maria-Irina Nicolae, Mathieu Sinn, Minh~Ngoc Tran, Beat Buesser, Ambrish Rawat,
  Martin Wistuba, Valentina Zantedeschi, Nathalie Baracaldo, Bryant Chen, Heiko
  Ludwig, et~al. 2018.
\newblock Adversarial robustness toolbox v1. 0.0.
\newblock \emph{arXiv preprint arXiv:1807.01069}.

\bibitem[{Papernot et~al.(2018)Papernot, Faghri, Carlini, Goodfellow, Feinman,
  Kurakin, Xie, Sharma, Brown, Roy, Matyasko, Behzadan, Hambardzumyan, Zhang,
  Juang, Li, Sheatsley, Garg, Uesato, Gierke, Dong, Berthelot, Hendricks,
  Rauber, and Long}]{papernot2018cleverhans}
Nicolas Papernot, Fartash Faghri, Nicholas Carlini, Ian Goodfellow, Reuben
  Feinman, Alexey Kurakin, Cihang Xie, Yash Sharma, Tom Brown, Aurko Roy,
  Alexander Matyasko, Vahid Behzadan, Karen Hambardzumyan, Zhishuai Zhang,
  Yi-Lin Juang, Zhi Li, Ryan Sheatsley, Abhibhav Garg, Jonathan Uesato, Willi
  Gierke, Yinpeng Dong, David Berthelot, Paul Hendricks, Jonas Rauber, and
  Rujun Long. 2018.
\newblock Technical report on the cleverhans v2.1.0 adversarial examples
  library.
\newblock \emph{arXiv preprint arXiv:1610.00768}.

\bibitem[{Papernot et~al.(2016{\natexlab{a}})Papernot, McDaniel, Jha,
  Fredrikson, Celik, and Swami}]{papernot2016limitations}
Nicolas Papernot, Patrick McDaniel, Somesh Jha, Matt Fredrikson, Z~Berkay
  Celik, and Ananthram Swami. 2016{\natexlab{a}}.
\newblock The limitations of deep learning in adversarial settings.
\newblock In \emph{2016 IEEE European symposium on security and privacy
  (EuroS\&P)}. IEEE.

\bibitem[{Papernot et~al.(2016{\natexlab{b}})Papernot, McDaniel, Swami, and
  Harang}]{papernot2016crafting}
Nicolas Papernot, Patrick McDaniel, Ananthram Swami, and Richard Harang.
  2016{\natexlab{b}}.
\newblock Crafting adversarial input sequences for recurrent neural networks.
\newblock In \emph{Proceedings of MILCOM}.

\bibitem[{Papineni et~al.(2002)Papineni, Roukos, Ward, and
  Zhu}]{papineni2002bleu}
Kishore Papineni, Salim Roukos, Todd Ward, and Wei-Jing Zhu. 2002.
\newblock Bleu: a method for automatic evaluation of machine translation.
\newblock In \emph{Proceedings of ACL}.

\bibitem[{Radford et~al.(2019)Radford, Wu, Child, Luan, Amodei, and
  Sutskever}]{radford2019language}
Alec Radford, Jeffrey Wu, Rewon Child, David Luan, Dario Amodei, and Ilya
  Sutskever. 2019.
\newblock Language models are unsupervised multitask learners.
\newblock \emph{OpenAI Blog}, 1(8).

\bibitem[{Rauber et~al.(2017)Rauber, Brendel, and Bethge}]{rauber2017foolbox}
Jonas Rauber, Wieland Brendel, and Matthias Bethge. 2017.
\newblock Foolbox: A python toolbox to benchmark the robustness of machine
  learning models.
\newblock In \emph{ICML Reliable Machine Learning in the Wild Workshop}.

\bibitem[{Reimers and Gurevych(2019)}]{reimers2019sentence}
Nils Reimers and Iryna Gurevych. 2019.
\newblock Sentence-bert: Sentence embeddings using siamese bert-networks.
\newblock In \emph{Proceedings of EMNLP}.

\bibitem[{Ren et~al.(2019)Ren, Deng, He, and Che}]{ren2019generating}
Shuhuai Ren, Yihe Deng, Kun He, and Wanxiang Che. 2019.
\newblock Generating natural language adversarial examples through probability
  weighted word saliency.
\newblock In \emph{Proceedings of ACL}.

\bibitem[{Ribeiro et~al.(2018)Ribeiro, Singh, and
  Guestrin}]{ribeiro2018semantically}
Marco~Tulio Ribeiro, Sameer Singh, and Carlos Guestrin. 2018.
\newblock Semantically equivalent adversarial rules for debugging {NLP} models.
\newblock In \emph{Proceedings of ACL}.

\bibitem[{Socher et~al.(2013)Socher, Perelygin, Wu, Chuang, Manning, Ng, and
  Potts}]{socher2013recursive}
Richard Socher, Alex Perelygin, Jean Wu, Jason Chuang, Christopher~D Manning,
  Andrew Ng, and Christopher Potts. 2013.
\newblock Recursive deep models for semantic compositionality over a sentiment
  treebank.
\newblock In \emph{Proceedings of EMNLP}.

\bibitem[{Szegedy et~al.(2014)Szegedy, Zaremba, Sutskever, Bruna, Erhan,
  Goodfellow, and Fergus}]{szegedy2014intriguing}
Christian Szegedy, Wojciech Zaremba, Ilya Sutskever, Joan Bruna, Dumitru Erhan,
  Ian Goodfellow, and Rob Fergus. 2014.
\newblock Intriguing properties of neural networks.
\newblock In \emph{Proceedings of ICLR}.

\bibitem[{Wallace et~al.(2019{\natexlab{a}})Wallace, Feng, Kandpal, Gardner,
  and Singh}]{wallace2019universal}
Eric Wallace, Shi Feng, Nikhil Kandpal, Matt Gardner, and Sameer Singh.
  2019{\natexlab{a}}.
\newblock Universal adversarial triggers for attacking and analyzing {NLP}.
\newblock In \emph{Proceedings of EMNLP-IJCNLP}.

\bibitem[{Wallace et~al.(2019{\natexlab{b}})Wallace, Tuyls, Wang, Subramanian,
  Gardner, and Singh}]{wallace2019allennlp}
Eric Wallace, Jens Tuyls, Junlin Wang, Sanjay Subramanian, Matt Gardner, and
  Sameer Singh. 2019{\natexlab{b}}.
\newblock Allennlp interpret: A framework for explaining predictions of nlp
  models.
\newblock In \emph{Proceedings of EMNLP-IJCNLP}.

\bibitem[{Zang et~al.(2020)Zang, Qi, Yang, Liu, Zhang, Liu, and
  Sun}]{zang2020word}
Yuan Zang, Fanchao Qi, Chenghao Yang, Zhiyuan Liu, Meng Zhang, Qun Liu, and
  Maosong Sun. 2020.
\newblock Word-level textual adversarial attacking as combinatorial
  optimization.
\newblock In \emph{Proceedings of ACL}.

\bibitem[{Zhang et~al.(2019)Zhang, Zhou, Miao, and Li}]{zhang2019generating}
Huangzhao Zhang, Hao Zhou, Ning Miao, and Lei Li. 2019.
\newblock Generating fluent adversarial examples for natural languages.
\newblock In \emph{Proceedings of ACL}.

\bibitem[{Zhang et~al.(2020)Zhang, Sheng, Alhazmi, and
  Li}]{zhang2020adversarial}
Wei~Emma Zhang, Quan~Z Sheng, Ahoud Alhazmi, and Chenliang Li. 2020.
\newblock Adversarial attacks on deep-learning models in natural language
  processing: A survey.
\newblock \emph{ACM Transactions on Intelligent Systems and Technology (TIST)},
  11(3):1--41.

\bibitem[{Zhao et~al.(2018)Zhao, Dua, and Singh}]{zhao2018generating}
Zhengli Zhao, Dheeru Dua, and Sameer Singh. 2018.
\newblock Generating natural adversarial examples.
\newblock In \emph{Proceedings of ICLR}.

\end{thebibliography}


\end{document}